\documentclass[graybox]{svmult}

\usepackage{type1cm}        % activate if the above 3 fonts are
\usepackage{makeidx}         % allows index generation
\usepackage{graphicx}        % standard LaTeX graphics tool
\usepackage{multicol}        % used for the two-column index
\usepackage[bottom]{footmisc}% places footnotes at page bottom

\usepackage{newtxtext}       % 
\usepackage[varvw]{newtxmath}       % selects Times Roman as basic font
\usepackage{algorithm}
\usepackage{algorithmic}
\usepackage{newfloat}
\usepackage{listings}
\usepackage{subcaption}
\usepackage{multirow}
\usepackage{amsmath}

\usepackage[backend=bibtex,url=false,style=numeric,maxnames=1]{biblatex}

\AtEveryBibitem{% Clean up the bibtex rather than editing it
 \clearfield{eprint}
 \clearfield{doi}
}
\makeindex             % used for the subject index
\begin{document}

\title*{Unsupervised Numerical Reasoning to Extract Phenotypes from Clinical Text by Leveraging External Knowledge}
\titlerunning{Unsupervised Numerical Reasoning to Extract Phenotypes from Clinical Text}
\author{Ashwani Tanwar, Jingqing Zhang, Julia Ive, Vibhor Gupta, Yike Guo}
\institute{
Ashwani Tanwar (\url{atanwar@pangaeadata.ai}), Jingqing Zhang (\url{jzhang@pangaeadata.ai}), Vibhor Gupta (\url{vgupta@pangaeadata.ai}) and Yike Guo \at Pangaea Data Limited, UK, USA
\and Jingqing Zhang and Yike Guo (\url{y.guo@imperial.ac.uk}) \at Data Science Institute, Imperial College London, UK
\and Julia Ive (\url{j.ive@qmul.ac.uk}) \at Department of Computing, Imperial College London, UK \at Queen Mary University of London, UK
\and Yike Guo \at Hong Kong Baptist University, Hong Kong SAR, China
\and Equal contribution: Ashwani Tanwar and Jingqing Zhang \at Corresponding to Yike Guo
}
\maketitle

\abstract*{Extracting phenotypes from clinical text has been shown to be useful for a variety of clinical use cases such as identifying patients with rare diseases. However, reasoning with numerical values remains challenging for phenotyping in clinical text, for example, temperature 102F representing Fever. Current state-of-the-art phenotyping models are able to detect general phenotypes, but perform poorly when they detect phenotypes requiring numerical reasoning. We present a novel unsupervised methodology leveraging external knowledge and contextualized word embeddings from ClinicalBERT for numerical reasoning in a variety of phenotypic contexts. Comparing against unsupervised benchmarks, it shows a substantial performance improvement with absolute gains on generalized Recall and F1 scores up to 79\% and 71\%, respectively. In the supervised setting, it also surpasses the performance of alternative approaches with absolute gains on generalized Recall and F1 scores up to 70\% and 44\%, respectively.}

\abstract{Extracting phenotypes from clinical text has been shown to be useful for a variety of clinical use cases such as identifying patients with rare diseases. However, reasoning with numerical values remains challenging for phenotyping in clinical text, for example, temperature 102F representing Fever. Current state-of-the-art phenotyping models are able to detect general phenotypes, but perform poorly when they detect phenotypes requiring numerical reasoning. We present a novel unsupervised methodology leveraging external knowledge and contextualized word embeddings from ClinicalBERT for numerical reasoning in a variety of phenotypic contexts. Comparing against unsupervised benchmarks, it shows a substantial performance improvement with absolute gains on generalized Recall and F1 scores up to 79\% and 71\%, respectively. In the supervised setting, it also surpasses the performance of alternative approaches with absolute gains on generalized Recall and F1 scores up to 70\% and 44\%, respectively.}


\section{Introduction}
\label{sec:intro}

Extracting phenotypes \footnote{In the medical text, the word ``phenotype'' refers to deviations from normal morphology, physiology, or behaviour, such as skin rash, hypoxemia, neoplasm, etc.~\cite{robinson2012deep}. Please note the difference of the phenotypic information to the diagnosis information expressed in ICD-10 codes \cite{10665-42980} as the former contributes to the latter.} from clinical text has been shown crucial for many clinical use cases \cite{zhang2021clinical} such as ICU in-hospital mortality prediction, remaining length of stay prediction, decompensation prediction and identifying patients with rare diseases. There are several challenges in extracting phenotypes such as handling a wide variety of phenotypic contexts, ambiguities, long term dependencies between phenotypes, and so on. Numerical reasoning is one of the key challenges as many of the phenotypes rely on bedside measurements such as temperature, blood pressure, heart rate, breathing rate, serum creatinine, hematocrit, glucose levels. We call these terms \textbf{numeric entities}. As these phenotypes require deep reasoning with the numbers, they are often missed or incorrectly predicted by the existing phenotype extraction methods \cite{Aronson2010,Savova2010,jonquet2009ncbo,deisseroth2019clinphen,arbabi2019ncr,kraljevic2019medcat,Tiwari2020}.

Existing phenotype extraction methods such as Neural Concept Recognizer (NCR) \cite{arbabi2019ncr} which are mostly based on state-of-the-art (SOTA) machine learning (ML) and natural language processing (NLP) technologies exploit non-contextualized word embeddings. These methods cannot detect contextual synonyms of the phenotypes which can be mentioned in various ways by clinicians. For example, previous SOTA phenotyping models like NCR and NCBO \cite{jonquet2009ncbo} can capture the phenotype \textit{Fever} from the sentence \textit{``patient is detected with fever''} but fail to capture the same from the sentence \textit{``patient is reported to have high temperature''}. Similarly, in the sentence, \textit{``patient is reported to have high temperature in the room with low temperature''}, only the former instance of \textit{temperature} is a phenotype, while the latter is not. The recent study \cite{zhang2021selfsupervised} demonstrates the capability of contextualized embeddings (BERT-based \cite{DBLP:conf/naacl/DevlinCLT19}) to differentiate the two instances by context. However, none of these methods above are specifically designed to reason with numbers in clinical text, for example, \textit{``temperature 102F''} representing \textit{Fever}. While the contextualized embeddings are useful for reasoning in different contexts, they are not sufficient to address numerical reasoning. 

In practice, the numerical reasoning for clinical context has specific challenges. First, clinical text may have accumulation of multiple numeric examples in a condensed context such as \textit{``Physical examination: temperature 97.5, blood pressure 124/55, pulse 79, respirations 18, O2 saturation 99\% on room air.''}. In addition, the numeric examples can be mentioned in a variety of different contexts such as \textit{``temperature of 102F''}, \textit{``temperature is 102F''}, \textit{``temperature is recorded as 102F''}, \textit{``temperature is found to be 102F''}, which make it more challenging to identify the \textit{(numeric entity, number)} pair namely \textit{(temperature, 102F)} in this case. Moreover, numbers in clinical text may not be necessarily connected with phenotypes. For example, the number in \textit{``patient required 4 days of hospitalization''} is not relevant to any phenotype.

To the best of our knowledge, previous studies have not addressed these challenges and we propose the first deep learning based (BERT-based) unsupervised methodology in this paper to accurately extract phenotypes by numerical reasoning from various contexts using external knowledge for clinical natural language processing. In summary, our \textbf{main contributions} are as follows:
\begin{enumerate}
    \item We propose a new approach to accurately detect phenotypes requiring numerical reasoning  using natural language processing and deep learning techniques.
    \item The approach is unsupervised and does not require manual data labelling. 
    \item Our approach can detect phenotypes from a variety of different contexts as it uses contextualized word embeddings.
    % \item The approach is generalisable to other similar problems which require numerical reasoning in the clinical textual notes. 
\end{enumerate}
\section{Related Work}
\textbf{Phenotyping:} Extraction of phenotypes from text has been addressed using several strategies. Shallow matching using linguistic patterns was used extensively by cTAKES \cite{Savova2010}, MetaMap \cite{Aronson2010}, and Clinphen \cite{deisseroth2019clinphen}. Then, the shallow matching was extended to semantic analysis by leveraging non-contextualized word embeddings by the works of \cite{arbabi2019ncr, kraljevic2019medcat,Tiwari2020}. For example, Neural Concept Recognizer (NCR) \cite{arbabi2019ncr} uses a convolutional neural network (CNN) to build non-contextualized embeddings by leveraging hierarchical medical concepts from biomedical ontologies like Human Phenotype Ontology (HPO) \cite{DBLP:journals/nar/0001GMCLVDBBBCC21}. Finally, \cite{zhang2021selfsupervised} showed that using contextualized embeddings from ClinicalBERT \cite{alsentzer-etal-2019-publicly} helps to detect contextual synonyms of the phenotypes from the text. Similarly, some other works \cite{liu-etal-2019-two, Yang2020, franz2020deep, li2020behrt} also exploited ClinicalBERT or BioBERT \cite{biobert} for phenotype detection, but all of these works focus on a limited set or group of phenotypes. None of these methods addresses phenotyping requiring numerical reasoning, so we extend the work and leverage external knowledge with ClinicalBERT to extract phenotypes requiring deep numerical reasoning in different textual contexts.
\\
\\
\noindent\textbf{Numerical Reasoning:} Recent works publish new datasets for numerical reasoning \cite{zhang-etal-2021-noahqa-numerical} and utilise deep learning based models to develop the numerical reasoning skills \cite{thawani-etal-2021-numeracy,duan-etal-2021-learning-numeracy,DBLP:journals/corr/abs-2101-11802,DBLP:journals/corr/abs-2109-03137} in respective domains other than the clinical domain. For example, \cite{DBLP:conf/acl/GevaGB20} shows gains by using artificially created data on various tasks involving numeracy such as math word problems and reading comprehension. Other works \cite{DBLP:conf/naacl/DuaWDSS019, DBLP:conf/emnlp/HuPHL19, DBLP:conf/emnlp/RanLLZL19} designed special modules for numerical reasoning in text which were then integrated with neural networks. Overall, these models have shown advancements in the respective domains for specialized problems but they did not incorporate clinical knowledge with specific extensive reasoning for clinical applications \cite{sushil-etal-2021-yet}. 
\section{Methodology}
\label{sec:methodology}

\begin{figure}[htb]
    \centering
    \includegraphics[width=0.8\textwidth]{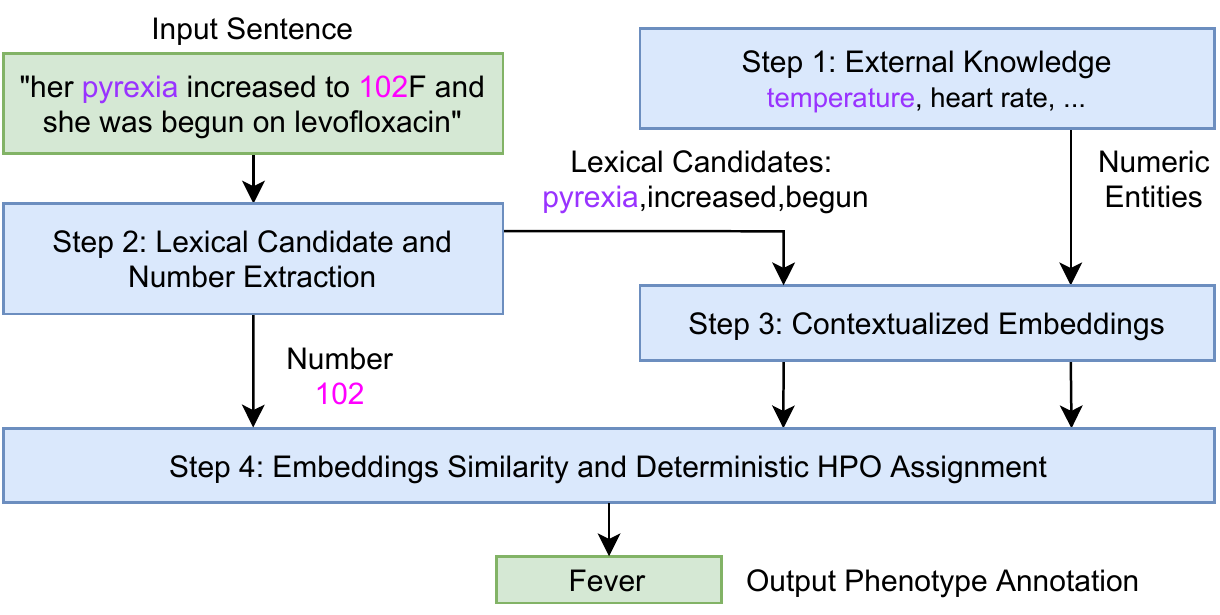}
    \caption{The workflow of the proposed Numerical Reasoning (NR) model to extract phenotypes from clinical text by leveraging external knowledge. The external knowledge is first created at one time as shown in Table~\ref{tab:external_knowledge_normal_reference_range_short} and Table \ref{tab:external_knowledge_numeric_entities_short}. Second, the numbers and lexical candidates are extracted from the input sentence. Third, the corresponding contextualized embeddings of lexical candidates and numeric entities are computed. As the final step, the contextualized embeddings are compared with similarity to find the closest numeric entity and then the phenotype (i.e., HPO concept) is assigned and annotated deterministically given the chosen numeric entity and the extracted number. All steps are further elaborated in Section \ref{sec:methodology}.
    }
    \label{fig:methodology}
\end{figure}

This section presents our unsupervised method for numerical reasoning (NR) in clinical textual notes to extract phenotypes. Figure~\ref{fig:methodology} shows the architecture of the proposed method which includes four steps (1) one-time creation of external knowledge connecting numeric entities and phenotypes, (2) extraction of numbers and lexical candidates for numeric entities from text, (3) creation of contextualized embeddings for numeric entities and lexical candidates and (4) linking candidates to numeric entities by embedding similarity and then determining corresponding phenotypes. This section elaborates all of the above steps.

\subsection{External Knowledge}
\label{ssec:methodology_external_knowledge}

% \usepackage{multirow}
\begin{table*}[h]
\centering
\scalebox{0.8}{
\begin{tabular}{|c|c|c|c|c|c|}
\hline
\multirow{2}{*}{\textbf{ID}} & \multirow{2}{*}{\textbf{Numeric Entity}} & \multirow{2}{*}{\textbf{Abbreviation}} & \multirow{2}{*}{\textbf{Unit}} & \multicolumn{2}{c|}{\textbf{Normal Reference Range}} \\ \cline{5-6} 
                             &                                          &                                &                                & \textbf{Lower Bound}            & \textbf{Upper Bound}           \\ \hline
0                            & temperature                              & temp                           & celsius                        & 36.4                      & 37.3                     \\ \hline
0                            & temperature                              & temp                           & fahrenheit                     & 97.5                      & 99.1                     \\ \hline
1                            & heart rate                               & heart rate                     & beats per minute (bpm)         & 60                        & 80                       \\ \hline
2                            & breathing rate                           & breathing rate                 & breaths per minute             & 12                        & 20                       \\ \hline
3                            & serum creatinine                         & serum creatinine               & mg/dL                          & 0.6                       & 1.2                      \\ \hline
3                            & serum creatinine                         & serum creatinine               & micromoles/L                   & 53                        & 106.1                    \\ \hline
4                            & hematocrit                               & hct                            & \%                             & 41                        & 48                       \\ \hline
5                            & blood oxygen                             & o2                             & \%                             & 95                        & 100                      \\ \hline
\end{tabular}
}
\caption{Examples of numeric entities that are used in the study with normal reference range and units. For example, the normal range of body temperature is 36.4 to 37.3 in Celsius or 97.5 to 99.1 in Fahrenheit. The ID column corresponds to that in Table~\ref{tab:external_knowledge_numeric_entities_short}.}
\label{tab:external_knowledge_normal_reference_range_short}
\end{table*}
% \usepackage{multirow}
\begin{table*}[h]
\centering
\scalebox{0.70}{
\begin{tabular}{|c|c|c|c|c|c|c|c|c|}
\hline
\multirow{2}{*}{\textbf{ID}} & \multirow{2}{*}{\textbf{\begin{tabular}[c]{@{}c@{}}Numeric \\ Entity\end{tabular}}} & \multirow{2}{*}{\textbf{Abb.}}                              & \multicolumn{2}{c|}{\textbf{\begin{tabular}[c]{@{}c@{}}Number Lower Than the Lower \\ Bound (Affirmed)\end{tabular}}} & \multicolumn{2}{c|}{\textbf{\begin{tabular}[c]{@{}c@{}}Number Higher than the Upper \\ Bound (Affirmed)\end{tabular}}} & \multicolumn{2}{c|}{\textbf{\begin{tabular}[c]{@{}c@{}}Number Inside Normal Range\\ (Negated)\end{tabular}}}      \\ \cline{4-9} 
                             &                                                                                     &                                                             & \textbf{HPO ID}       & \textbf{HPO Name}                                                                               & \textbf{HPO ID}    & \textbf{HPO Name}                                                                                  & \textbf{HPO ID} & \textbf{HPO Name}                                                                                \\ \hline
0                            & temperature                                                                         & temp                                                        & HP:0002045            & Hypothermia                                                                                     & HP:0001945         & Fever                                                                                              & HP:0004370      & \begin{tabular}[c]{@{}c@{}}Abnormality of \\ temperature regulation\end{tabular}                 \\ \hline
1                            & heart rate                                                                          & heart rate                                                  & HP:0001662            & Bradycardia                                                                                     & HP:0001649         & Tachycardia                                                                                        & HP:0011675      & Arrhythmia                                                                                       \\ \hline
2                            & breathing rate                                                                      & \begin{tabular}[c]{@{}c@{}}breathing \\ rate\end{tabular}   & HP:0046507            & Bradypnea                                                                                       & HP:0002789         & Tachypnea                                                                                          & HP:0002793      & \begin{tabular}[c]{@{}c@{}}Abnormal pattern of \\ respiration\end{tabular}                       \\ \hline
3                            & serum creatinine                                                                    & \begin{tabular}[c]{@{}c@{}}serum \\ creatinine\end{tabular} & HP:0012101            & \begin{tabular}[c]{@{}c@{}}Decreased \\ serum creatinine\end{tabular}                           & HP:0003259         & \begin{tabular}[c]{@{}c@{}}Elevated \\ serum creatinine\end{tabular}                               & HP:0012100      & \begin{tabular}[c]{@{}c@{}}Abnormal circulating \\ creatinine concentration\end{tabular}         \\ \hline
4                            & hematocrit                                                                          & hct                                                         & HP:0031851            & Reduced hematocrit                                                                              & HP:0001899         & Increased hematocrit                                                                               & HP:0031850      & Abnormal hematocrit                                                                              \\ \hline
5                            & blood oxygen                                                                        & o2                                                          & HP:0012418            & Hypoxemia                                                                                       & HP:0012419         & Hyperoxemia                                                                                        & HP:0500165      & \begin{tabular}[c]{@{}c@{}}Abnormal blood \\ oxygen level\end{tabular}                           \\ \hline
\end{tabular}
}
\caption{Examples of numeric entities that are used in the study with phenotype labels (including HPO ID and HPO name). Each numeric entity is connected with three phenotype concepts. For example, a patient has \textit{Hypothermia (Fever)} if their body temperature is lower (higher) than the lower (upper) normal limit. If the body temperature of the patient is inside the normal range, it is negation of the general phenotype \textit{Abnormality of temperature regulation}. The ID column corresponds to Table~\ref{tab:external_knowledge_normal_reference_range_short}.}
\label{tab:external_knowledge_numeric_entities_short}
\end{table*}

Phenotypes can often be inferred from ranges of numerical values together with numeric entities which are mentioned in clinical text. For example, the clinicians may mention the numeric entity \textit{temperature} and the value ``\textit{102 Fahrenheit}'' in clinical notes to suggest a patient is suffering from the phenotype \textit{Fever (HP:0001945)}. Therefore, we create an external knowledge base to formalise such connections between phenotypes, numeric entities and numerical values.

We first manually collect a list of 33 most frequent numeric entities such as temperature, heart rate, breathing rate, serum anion gap and platelet and their corresponding normal reference ranges (values and units) from the website of National Health Service of UK \footnote{Accessed in November 2021: https://www.nhs.uk} and MIMIC-III database \cite{johnson2016mimic}. Table~\ref{tab:external_knowledge_normal_reference_range_short} shows examples of numeric entities and their corresponding lower/upper bounds with units. For example, the normal body temperature has a lower bound 97.5 Fahrenheit (36.4 Celsius) and an upper bound 99.1 Fahrenheit (37.3 Celsius).

Those numeric entities are then manually mapped with phenotypes which are defined and standardised by \textbf{Human Phenotype Ontology (HPO)} \cite{10.1093/nar/gkw1039}. In most cases, a numeric entity corresponds to three phenotypes depending on whether the actual measurement is lower than, higher than or within the normal reference range. If the actual measurement is lower (higher) than the lower (upper) bound, it means the relevant phenotype is affirmed. For example, the phenotype \textit{Hypothermia (HP:0002045)} or \textit{Fever (HP:0001945)} is affirmed if the body temperature is lower or higher than the normal range, respectively. Otherwise, if the body temperature is inside the normal range, the general phenotype \textit{Abnormality of Temperature Regulation (HP:0004370)}, which is the parent phenotype of \textit{Hypothermia (HP:0002045)} or \textit{Fever (HP:0001945)}, is negated. Table~\ref{tab:external_knowledge_numeric_entities_short} demonstrates examples of the connections between numeric entities and phenotypes. 

Both Table~\ref{tab:external_knowledge_normal_reference_range_short} and Table~\ref{tab:external_knowledge_numeric_entities_short} are validated by three expert clinicians with consensus for authenticity and consistency. The external knowledge 
% and corresponding contextualized embeddings of numeric entities in Section~\ref{ssec:methodology_sts_contextual_embeddings} 
is created at one time and prior to other steps, which makes the external knowledge reusable.

% Phenotype for \textit{inside the range} is used to \textit{negate} the other 2 phenotypes, and thus marked with \textit{NEGATION} category. Contrary, other 2 phenotypes are marked with \textit{AFFIRMED} category. For example, as shown in the above figure, (HP:0002045,Hypothermia), (HP:0004370,Abnormality of temperature regulation), and (HP:0001945,Fever) are mapped with the temperature when the measurement is \textit{less than the normal lower limit}, \textit{inside the normal range}, and \textit{more than the normal upper limit}, respectively. A list of all the numeric entities along with their mapped phenotypes is shown in Table~\ref{tab:external_knowledge_numeric_entities_short}. Both of the above tables were further validated by expert clinicians for authenticity and consistency.

\subsection{Number and Lexical Candidates Extraction}
\label{ssec:methodology_number_entities_candidates}

\begin{figure*}[htb]
    \centering
    \includegraphics[scale=0.5]{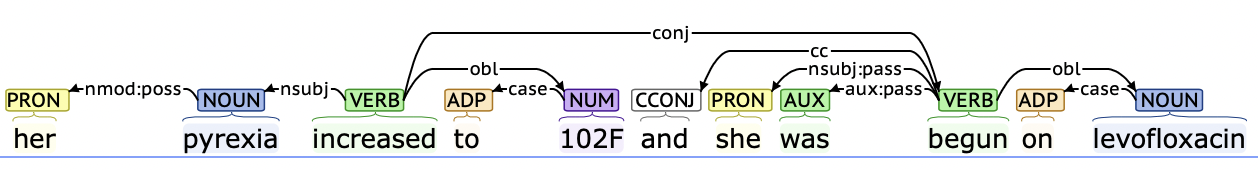}
    \caption{An example of syntactic analysis to extract lexical candidates. In this example, we extract \textit{pyrexia, increased, begun} as the lexical candidates from the sentence by using Part-of-Speech (POS) tagging and dependency parsing. 
    }
    \label{fig:methodology_syntactic_analysis}
\end{figure*}

We then extract \textbf{numbers} and their corresponding \textbf{lexical candidates} which are likely to be numeric entities from clinical text. For example, in the input sentence ``her \textit{pyrexia} increased to \textit{102F} and she was begun on levofloxacin'', the number \textit{``102''} is connected with \textit{``pyrexia''} which is contextually similar to the numeric entity ``temperature''. The regular expression patterns are created to extract the numbers which typically appear in alpha-numeric format such as ``pyrexia increased to \textit{102F}'' and ``heart rate in \textit{90s}''. The numbers which are dates or part of specific clinical concepts such as ``vitamin B\textit{12}'', ``O\textit{2} saturation'' are excluded by using a pre-defined dictionary of alpha-numeric words \cite{DBLP:journals/jamia/MoonPLRM14} as the numbers are not relevant to phenotypes.

After the extraction of numbers, the lexical candidates connected to these numbers are extracted using syntactic analysis. As shown in Figure \ref{fig:methodology_syntactic_analysis},
% We study syntactic relations between a number and its corresponding numeric entity in roughly 20 sentences to design Algorithm~\ref{alg:cand_extraction}. 
we focus on (proper) nouns, adjectives and verbs 
% use only the words with the Universal Dependencies \cite{DBLP:conf/lrec/NivreMGHMPSTZ20} 
that are connected with syntactic connections to the extracted numbers (heads or children in the syntactic tree). As a special case, we allow one additional hop from the extracted number via the dependency relation `obl' which stands for oblique nominal. For example, in Figure \ref{fig:methodology_syntactic_analysis}, the words \textit{``pyrexia''}, \textit{``increased''}, and \textit{``begun''} are extracted as lexical candidates because they are connected to the extracted number \textit{``102F''} and therefore are likely to represent numeric entities.   

The list of extracted lexical candidates is passed to the following steps to decide the corresponding numeric entities based on context. As the extraction method of lexical candidates is designed to encourage more extraction to increase recall so that no important word is missed, not all of the lexical candidates will eventually correspond to a numeric entity.

% At this moment, we have a number which \textit{can be} a part of a phenotype and a list of lexical candidates which \textit{can be} numeric entities related to the number. In Section~\ref{ssec:methodology_relation_prediction}, an exact word is identified from these lexical candidates which is numeric entity of the number.

\subsection{Contextualized Embeddings for Numeric Entities and Lexical Candidates }
\label{ssec:methodology_sts_contextual_embeddings}

% \red{JZ comment: use equation, diagram and psedo-code to avoid tedious description.}

% \red{JZ comment: the Methodology section should focus on the FINAL proposed method. The baselines such as pre-traing embeddings without fine-tuning should be moved to the Results section. The limitation and future improvement can be moved to Conclusion and Future Work section.}

We use contextualized embeddings (ClinicalBERT \cite{alsentzer-etal-2019-publicly}) of numeric entities and lexical candidates to measure their similarity and decide which numeric entity should be assigned to the extracted lexical candidates from the input sentence.  

% of all numeric entities upfront (which are listed in Table \ref{tab:external_knowledge_numeric_entities_short}) and the extracted lexical candidates on-the-fly.

The objective of the model is to learn a semantic space where all possible expressions (including names and synonyms) of one numeric entity are clustered while the expressions of different numeric entities are differentiated. 
To achieve this, we use ClinicalBERT finetuned with Semantic Textual Similarity (STS) objective defined as follows:

\begin{equation}
\begin{aligned}
\label{eqn:finetuning_loss}
\mathcal{L}~(e_i, s_j) = \frac{1}{|\mathcal{E}|}\frac{1}{|\mathcal{S}|}\sum_{i=1}^{|\mathcal{E}|}\sum_{j=1}^{|\mathcal{S}|}\bigg(\cos(\vec{h}_{e_i},\vec{h}_{s_j}) - y_{e_i,s_j}\bigg)^2,\\ \text{where}~~y_{e_i,s_j} = \begin{cases}
			1, & \text{if $s_j$ is a synonym of $e_i$}\\
            0, & \text{otherwise}
		 \end{cases} 
\end{aligned}
\end{equation}

\noindent where $\vec{h}_{e_i}$ represents contextualized embedding for the $i^{\text{th}}$ numeric entity $e_i$ in $\mathcal{E}$. Similarly, $\vec{h}_{s_j}$ represents contextualized embedding for the $j^{\text{th}}$ synonym $s_j$ in $\mathcal{S}$. The ground truth label $y_{e_i,s_j}$ is 1 if the synonym $s_j$ is one of the synonyms of the numeric entity $e_i$ and 0 if otherwise. The loss function aims to maximise the cosine similarity between numeric entities and their corresponding synonyms and minimise the similarity between numeric entities and irrelevant synonyms. The collection of $\{\vec{h}_{e_i} | \forall e_i \in \mathcal{E} \}$ is used as the reference contextualized embeddings of numeric entities created once. 

As the training data, we collect all synonyms $\mathcal{S} = \{s_1, s_2, \dots, s_{|\mathcal{S}|} \}$ of all the numeric entities $\mathcal{E} = \{e_1, e_2, \dots, e_{|\mathcal{E}|} \}$ (listed in Table~\ref{tab:external_knowledge_numeric_entities_short}) by connecting the HPO IDs with Unified Medical Language System (UMLS) \cite{DBLP:journals/nar/Bodenreider04}.

During inference, lexical candidates extracted from input sentences are fed into the finetuned ClinicalBERT based model to produce their contextualized embeddings. 

% Further, negative sampling is used where the cosine similarity is minimized between the numeric entities and the UMLS synonyms of \textit{other} entities.
% In Section~\ref{ssec:results_umap_visualization}, a comparative analysis between pretrained and fine-tuned embeddings is done. \red{JZ comment: what is the narrative of comparing these here? If not important here, please move to Result section.}

% \subsection{Relation Prediction between Number and Numeric Entities using Contextualized Embeddings}
\subsection{Embedding Similarity and Deterministic HPO Assignment}
\label{ssec:methodology_relation_prediction}

% \red{JZ comment: use equation, diagram and psedo-code to avoid tedious description.}

% \input{pseudo-codes/rel_prediction}

Embeddings pairs are formed by Cartesian product of the contextualized embeddings of lexical candidates and reference contextualized embeddings of numeric entities. Then cosine similarity is computed between all the pairs. The pair with the maximum cosine score above a pre-set threshold gives the selected lexical candidate which in turn gives the corresponding numeric entity. 

% After extracting all the lexical candidates for numeric entities, we use contextualized embeddings to identify the exact numeric entity connected with the number from these lexical candidates, as illustrated in Algorithm~\ref{alg:rel_extraction}. We build the contextualized embeddings for all the lexical candidates using the same \textit{fine-tuned} ClinicalBERT vector space introduced in Section~\ref{ssec:methodology_sts_contextual_embeddings}. Then, a Cartesian product is done between the contextualized embeddings of all the lexical candidates and the predefined numeric entities listed in Table~\ref{tab:external_knowledge_numeric_entities_short}. Then, cosine scores are calculated between all the pairs. A pair with the highest cosine score exceeding the threshold of 0.90 is highly likely to give a valid numeric entity. In this way, we determine a lexical candidate as a valid numeric entity which is contextually similar to one of the predefined entities in the external knowledge. By following this, we predict the numeric entity as well as implicitly map the predicted entity to one of the records in the external knowledge. 
A sentence may have multiple numbers connected with their corresponding numeric entities. We simply consider the lexical candidates (corresponding to each number) as an independent case for the above Cartesian product which helps extracting multiple candidate numeric entities from a single sentence.

After measuring similarity of embeddings and determining the numeric entities, we deterministically assign the phenotype depending if the corresponding number is lower than the lower bound, inside the normal range, or higher than the upper bound. For example, in Figure \ref{fig:methodology}, the lexical candidate ``pyrexia'' is extracted and the numeric entity ``temperature'' is assigned based on contextualized embedding. As the number ``102F'' is higher than the upper bound ``99.1'', the phenotype \textit{Fever (HP:0001945)} is eventually assigned.

% Thus, in the figure, we obtain Cartesian product between the embeddings of lexical candidates, i.e., (\textit{pyrexia}, \textit{increased}, and \textit{begun}) and the embeddings of all the numeric entities from Table~\ref{tab:external_knowledge_numeric_entities_short}. Then, we compare the cosine scores among all the pairs and find the pair \textit{(pyrexia, temperature)} having the highest score. So, we pick the record from the external knowledge corresponding to the numeric entity \textit{temperature} in Table~\ref{tab:external_knowledge_numeric_entities_short}. Then, we predict the phenotype \textit{(HP:0001945, Fever)} deterministically after finding the number \textit{102} higher than the upper limit of the normal reference range in Table~\ref{tab:external_knowledge_normal_reference_range_short}.   

% \subsection{Minor Optimizations}

% \red{JZ comment: this section can be merged into previous section as a new paragraph}

We also enhance the HPO assignment process by handling different units of numbers (e.g. Fahrenheit and Celsius) because sometimes the units are not explicitly mentioned in text. 
% \red{JZ comment: the word ``some'' and ``such as'' here may raise concern of reproducibility. Summarise and describe them all confidently.}. 
Therefore, we decide the unit by comparing the ratio of the number to the extreme ends of the normal reference ranges in different units. For example, normal range for temperature is (36.4, 37.3) in Celsius and (97.5, 99.1) in Fahrenheit. If a given number is 92, then we take the ratios as the following. The unit giving the smaller ratio (Fahrenheit in this case) is then used to determine HPO assignment. 
\begin{multicols}{2}
\begin{itemize}
    \item{\(\displaystyle \frac{\text{number}}{\text{upper\_bound\_celsius}}\) = \(\displaystyle \frac{92}{37.3}\) = 2.5}
    \item{\(\displaystyle \frac{\text{lower\_bound\_fahren}}{\text{number}}\) = \(\displaystyle \frac{97.5}{92}\) = 1.1}
\end{itemize}
\end{multicols}
% \usepackage{multirow}
\begin{table*}[h]
\centering
\scalebox{0.80}{
\renewcommand{\arraystretch}{1.3}
\begin{tabular}{|c|c|c|c|c|c|c|}
\hline
\multicolumn{2}{|c|}{\textbf{Primary Phenotype}} & \multirow{2}{*}{\textbf{Unit}} & \multicolumn{2}{c|}{\textbf{Granular Range}} & \multicolumn{2}{c|}{\textbf{Granular Phenotype}} \\ \cline{1-2} \cline{4-7} 
\textbf{HPO ID}       & \textbf{HPO Name}                  &                                & \textbf{Lower}        & \textbf{Upper}       & \textbf{HPO ID} & \textbf{HPO Name}                        \\ \hline
HP:0012664        & Reduced ejection fraction      & \%                             & 0                     & 29.9                 & HP:0012666  & Severely reduced ejection fraction   \\ \hline
HP:0012664        & Reduced ejection fraction      & \%                             & 30                    & 39.9                 & HP:0012665  & Moderately reduced ejection fraction \\ \hline
HP:0012664        & Reduced ejection fraction      & \%                             & 40                    & 49.9                 & HP:0012663  & Mildly reduced ejection fraction     \\ \hline
HP:0001945        & Fever                          & celsius                        & 37.4                  & 38                   & HP:0011134  & Low-grade fever                      \\ \hline
HP:0001945        & Fever                          & fahrenheit                     & 99.2                  & 100.4                & HP:0011134  & Low-grade fever                      \\ \hline
\end{tabular}}
\caption{A list of granular phenotypes under primary phenotypes. For example, the reduced ejection fraction can be further divided into three sub-phenotypes by severity based on the actual percentage mentioned in clinical text.}
\label{tab:external_knowledge_granular_hpo}
\end{table*}

Moreover, we consider granular phenotypes based on granular sub-ranges as shown in Table~\ref{tab:external_knowledge_granular_hpo}. 
% For example, reduced ejection fraction is divided into severely, moderately and mildly reduced ejection fraction depending on the numeric readings of ejection fraction.
\section{Experiment Design}
\label{sec:experiment_design}

\subsection{Datasets}
% \usepackage{multirow}
\begin{table*}[h]
\centering
\scalebox{0.95}{
\begin{tabular}{|c|c|c|c|c|c|}
\hline
% \multicolumn{6}{|c|}{Counts} \\ \hline
\multicolumn{3}{|c|}{\textbf{Test Set (Unsupervised Setting)}} & \multicolumn{3}{|c|}{\textbf{Test Set (Supervised Setting)}} \\ \hline
\textbf{EHRs} & \textbf{All phenotypes} & \textbf{NR-specific phenotypes} & \textbf{EHRs} & \textbf{All phenotypes} & \textbf{NR-specific phenotypes} \\ \hline
705   & 20926     & 1121      & 170   & 5047      & 322     \\ \hline
\end{tabular}
}
\caption{Statistics (counts) of the test sets in the unsupervised and supervised setting, respectively. The test set in the unsupervised setting includes all manually annotated EHRs. The test set in the supervised setting is a subset of that in the unsupervised setting because some annotated EHRs are used to finetune the baseline models. Please note only Numerical Reasoning (NR) specific phenotypes are used for evaluation as the other phenotypes are not related with numbers in clinical narratives. }
\label{tab:dataset_details}
\end{table*}

We use clinical textual notes from the publicly available MIMIC-III database \cite{johnson2016mimic}. In the unsupervised setting, we collected 705 EHR textual notes with 20,926 gold phenotype annotations as shown in Table~\ref{tab:dataset_details}. The gold phenotype annotations were created by three expert clinicians with consensus and the clinicians were specifically asked to identify contextual synonyms of phenotypes
such as ``drop in blood pressure'' and ``BP of 79/48'' for \textit{Hypotension (HP:0002615)}. Out of these phenotype annotations, we select a subset with 1,121 phenotype annotations (i.e., NR specific phenotypes) which require numerical reasoning based on two criteria: (1) the annotated phenotypes are among one of the HPO IDs that require numerical reasoning as mentioned in Table~\ref{tab:external_knowledge_numeric_entities_short} and Table \ref{tab:external_knowledge_granular_hpo} and (2) the corresponding textual spans of phenotypes contain numbers. The test set in the unsupervised setting is used to compare the proposed NR model with previous unsupervised baseline methods.

In the supervised setting, as 535 out of 705 manually annotated EHRs are used to finetune the baseline methods (like ClinicalBERT), the remaining 170 EHRs are used for testing. In other words, the test set in the supervised setting is the subset of that in the unsupervised setting. Though, the proposed NR model is strictly unsupervised, we compare it with supervised baselines to rigorously assess its performance.

% The datasets for extrinsic evaluation are also created based on MIMIC-III database and the details will be discussed in Section \ref{ssec:results_use_cases}.

\subsection{Implementation Details}
\label{subsec:implementation_details}
We use the Stanford Stanza \cite{qi2020stanza, 10.1093/jamia/ocab090} library to extract the lexical candidates for numeric entities using syntactic analysis. In syntactic analysis, we only focus on nouns, adjectives and verbs that are ``NOUN'', ``PROPN'', ``ADJ'', and ``VERB'' as marked by the Part of speech (POS) tagger and we also optimise the process by adding words with the dependency relation `compound' to capture multi-word phrases like ``heart rate'' and ``blood pressure''. Then, we use Semantic Textual Similarity (STS) model from Sentence Transformers \cite{DBLP:conf/emnlp/ReimersG19} library to finetune the ClinicalBERT embeddings with cosine similarity up to 4 epochs using their default hyperparameters\footnote{Accessed in November 2021: \url{https://www.sbert.net/docs/training/overview.html}} along with a train and validation batch size of 16 and 1000 evaluation steps. Mean pooling is used to get embeddings of multi-word UMLS synonyms. The threshold for embedding similarity is set as 0.9 empirically. The implementation of the proposed method also uses some other third-party libraries including PyTorch \cite{NEURIPS2019_bdbca288} and
% Pandas \cite{mckinney-proc-scipy-2010, jeff_reback_2021_5501881} and 
spaCy. 

\subsection{Baselines and Evaluation Methods}

We compare the proposed NR model with previous state-of-the-art phenotyping models. In the unsupervised setting, the proposed NR model is compared with unsupervised baselines including NCBO \cite{jonquet2009ncbo}, NCR \cite{arbabi2019ncr} and the unsupervised model by \cite{zhang2021selfsupervised}. In the supervised setting, the proposed NR model is compared with the finetuned ClinicalBERT \cite{alsentzer-etal-2019-publicly} (which is finetuned for phenotyping) and the supervised model by \cite{zhang2021selfsupervised}. The NCBO, NCR and finetuned ClinicalBERT are selected as they show better performance than other baseline phenotyping methods (including cTAKES \cite{Savova2010}, MetaMap \cite{Aronson2010}, Clinphen \cite{deisseroth2019clinphen}, MedCAT \cite{kraljevic2019medcat}, BERT \cite{DBLP:conf/naacl/DevlinCLT19}, BioBERT \cite{biobert}, SciBERT \cite{beltagy-etal-2019-scibert}) in corresponding settings as demonstrated by \cite{zhang2021selfsupervised}. Please note the work by \cite{zhang2021selfsupervised} publishes one unsupervised and one supervised model hence we compare the proposed NR model with both. We decide not to compare with recent numerical reasoning models (such as \cite{thawani-etal-2021-numeracy,duan-etal-2021-learning-numeracy,DBLP:journals/corr/abs-2101-11802,DBLP:journals/corr/abs-2109-03137}) as none of them incorporates clinical knowledge and we find it costly to adapt them to the clinical domain.

We first evaluate the proposed NR model against the baselines by using micro-averaged Precision, Recall and F1-score at the document level. To ensure comparison with previous studies, we follow the practice by \cite{Liu2019} and compute the metrics by the following two strategies. (1) Exact Matches: Only the exact same HPO annotations against the gold standard annotations are counted as correct; (2) Generalized Matches: the gold standard annotations as well as predicted HPO annotations are extended to include all ancestor HPO concepts until the root concept \textit{Phenotypic Abnormality (HP:0000118)} (exclusive) in the HPO hierarchy. All the extended HPO annotations are then de-duplicated and added to the list of gold standard and predicted HPO annotations respectively for evaluation. By the generalized matches, the prediction of HPO concepts which are children, neighbours or ancestors of the target HPO concepts also receives credits.

% For extrinsic evaluation, we report sensitivity and specificity of the proposed NR model on the downstream clinical use cases where we predict if patients are diagnosed by rare diseases, namely, Pulmonary Arterial Hypertension (PAH) and Lupus Nephritis, based on the phenotypic features of the patients. For the three ICU tasks, following the practice by \cite{harutyunyan2019multitask}, we report Kappa score and AUC-ROC.
\section{Results and Discussion}
\label{sec:results_and_discussion}

\subsection{Quantitative Analysis}
\label{ssec:results_quantitative}

We report our quantitative results in Table~\ref{tab:results_mimic_unsupervised} where we evaluate the NR model in the unsupervised setting. We also compare the NR model with the baselines -- NCBO, NCR, and unsupervised model by \cite{zhang2021selfsupervised} but they perform poorly on the unsupervised test set with straight 0 on all the metrics. This is expected as they are not designed to handle numbers. The NR model performs significantly better than all of them achieving 69\% recall and 59\% F1 using exact metrics, while 79\% recall and 71\% F1 using generalized metrics. Precision is relatively lower as we focus on recall to extract more phenotypes, which is motivated by the preference that a model is sensitive to capture more phenotypic features of patients rather than missing ones for better accuracy in downstream clinical use cases \cite{zhang2021clinical}. Overall, the NR model shows huge gains which is useful in the absence of costly annotated data. 

We also compare the unsupervised NR model with the previous state-of-the-art supervised baseline methods. First, we compare the NR model with the supervised model by \cite{zhang2021selfsupervised} which is finetuned with annotated data. This comparison is shown in Table~\ref{tab:results_mimic_supervised} on the supervised test set. Though the supervised model by \cite{zhang2021selfsupervised} outperforms its unsupervised version, the proposed unsupervised NR model performs better than the supervised baseline with gains of 12.5\% and 5.7\% on exact and generalized recall, respectively. However, there is a drop in precision which results in the comparable F1 scores. Moreover, using a combination of both the models achieves the best performance improving score by 21.5\% and 14.3\% on exact and generalized recall, respectively, and 4.3\% and 0.7\% gains on exact and generalized F1 scores, respectively. Then, the NR model is compared against the finetuned ClinicalBERT \cite{alsentzer-etal-2019-publicly} which is finetuned to detect phenotypes. The combination of NR model and supervised model by \cite{zhang2021selfsupervised} surpasses the performance of the baseline with gains of 66.4\% and 69.7\% on exact and generalized recall, respectively and 40\% and 44.2\% gains on exact and generalized F1 scores, respectively, as shown in Table~\ref{tab:results_mimic_supervised}. These results highlight the impact of the NR model which shows better performance than the supervised models eliminating the need of costly human annotations of phenotypes. 

% Moreover, the model assisted the clinicians to find the phenotypes requiring numerical reasoning initially missed by them.

% \usepackage{multirow}
\begin{table*}[]
\centering
\scalebox{0.99}{
\begin{tabular}{|c|c|c|c|c|c|c|}
\hline
\multirow{2}{*}{\textbf{Model}} & \multicolumn{3}{c|}{\textbf{Exact}}                & \multicolumn{3}{c|}{\textbf{Generalized}}          \\ \cline{2-7} 
                                & \textbf{Precision} & \textbf{Recall} & \textbf{F1} & \textbf{Precision} & \textbf{Recall} & \textbf{F1} \\ \hline
NCBO                            & 0                  & 0               & 0           & 0                  & 0               & 0           \\ \hline
NCR                             & 0                  & 0               & 0           & 0                  & 0               & 0           \\ \hline
\cite{zhang2021selfsupervised} (unsupervised)               & 0                  & 0               & 0           & 0                  & 0               & 0           \\ \hline
Numerical Reasoning (NR)                              & \textbf{0.5176}             & \textbf{0.6879}          & \textbf{0.5907}      & \textbf{0.6479}             & \textbf{0.7907}          & \textbf{0.7122}      \\ \hline
\end{tabular}
}
\caption{
In the unsupervised setting, the comparison of baselines NCBO, NCR, and \cite{zhang2021selfsupervised} (unsupervised) with proposed Numerical Reasoning (NR) model shows the superior performance of NR model. Interestingly but not surprisingly, the baseline methods produce zero accuracy as they are not designed to reason by numbers.}
\label{tab:results_mimic_unsupervised}
\end{table*}

% \usepackage{multirow}
\begin{table*}[h]
\centering
\scalebox{0.99}{
\begin{tabular}{|c|c|c|c|c|c|c|}
\hline
\multirow{2}{*}{\textbf{Model}} & \multicolumn{3}{c|}{\textbf{Exact}}                    & \multicolumn{3}{c|}{\textbf{Generalized}}              \\ \cline{2-7} 
                                & \textbf{Precision} & \textbf{Recall} & \textbf{F1}     & \textbf{Precision} & \textbf{Recall} & \textbf{F1}     \\ \hline
Finetuned ClinicalBERT                    & \textbf{0.8235}    & 0.181           & 0.2968          & \textbf{1.000}         & 0.2229          & 0.3646          \\ \hline
\cite{zhang2021selfsupervised} (supervised)                  & 0.6791             & 0.6293          & 0.6532          & 0.8245             & 0.7762          & 0.7996          \\ \hline
Numerical Reasoning (NR)                              & 0.5952             & 0.7543          & 0.6654          & 0.7290              & 0.8339          & 0.7780           \\ \hline
\cite{zhang2021selfsupervised} (supervised) + NR             & 0.5921             & \textbf{0.8448} & \textbf{0.6963} & 0.7175             & \textbf{0.9201} & \textbf{0.8062} \\ \hline
\end{tabular}
}
\caption{The comparison of supervised baselines with the proposed Numerical Reasoning (NR) model in the supervised setting shows that the NR model increases recall significantly by finding more phenotypes even without supervision. Please note supervised setting refers to a subset of unsupervised setting test set which is created to compare unsupervised NR with the supervised baselines.}
\label{tab:results_mimic_supervised}
\vspace{-9mm}
\end{table*}

\subsection{Qualitative Analysis}
\label{ssec:results_qualitative}

We investigate the numerical reasoning capabilities of the proposed NR model and other baseline methods by eye-balling example sentences having different contexts. In the sentence \textit{``patient has a temperature of 102F.''}, NCR, NCBO, and \cite{zhang2021selfsupervised} (unsupervised) do not detect any phenotype. But after adding the word \textit{high}, i.e., \textit{``patient has a \textbf{high} temperature of 102F.''}, \cite{zhang2021selfsupervised} (unsupervised) correctly detects the phenotype \textit{Fever (HP:0001945)}. However, the predicted textual span is \textit{``high temperature''} only ignoring the number \textit{102F}. It indicates that the \cite{zhang2021selfsupervised} (unsupervised) relies on context without considering numbers, while NCR and NCBO still do not detect any phenotype. When the word \textit{``temperature''} is changed to \textit{``fever''} and the whole sentence becomes \textit{``patient has a high \textbf{fever} of 102F.''}, all three unsupervised baseline methods can correctly detect the phenotype \textit{Fever (HP:0001945)} though the the number is still ignored in the predicted textual span. Overall, we observe all the unsupervised baseline methods solely rely on the textual content by ignoring the numbers, though \cite{zhang2021selfsupervised} (unsupervised) can find contextual synonyms of phenotypes. 

In contrast, the proposed NR model correctly detects the phenotype from all the three variants of the original sentence with the correct textual spans which include numbers. More precisely, the target textual spans are \textit{``temperature of 102F''}, \textit{``temperature of 102F''}, and \textit{``fever of 102F''} with the phenotype \textit{Fever (HP:0001945)} for the three sentences above, respectively. We observe the similar behavior given the sentence \textit{``patient has a breathing rate of 27."} with the phenotype \textit{Tachypnea (HP:0002789)} as well as \textit{``patient has a serum creatinine of 1.7."} with the phenotype \textit{Elevated serum creatinine (HP:0003259)}. The model \cite{zhang2021selfsupervised} (unsupervised) detects the phenotype (still ignoring the numbers) when an indicative word like \textit{``high''} is added, while NCBO and NCR miss the annotations with the exception for the latter sentence where NCR detects the phenotype after \textit{``high''} is added to the sentence. In short, the results suggest that the proposed NR model reasons with the numbers effectively in different contexts without supervision.

The supervised model by \cite{zhang2021selfsupervised} overall performs much better with reasonable accuracies than the unsupervised baselines which give straight 0 scores. However, it still lacks the capabilities to reason with numbers. For instance, though the \cite{zhang2021selfsupervised} (supervised) correctly predicts the phenotype \textit{Fever (HP:0001945)} from the sentence \textit{``patient has a temperature of 102F."}, if the number in the sentence is changed from 102F to 92F and the target phenotype is therefore changed to \textit{Hypothermia (HP:0002045)}, the \cite{zhang2021selfsupervised} (supervised) still predicts fever mistakenly. Similar incorrect predictions are observed when the target phenotype is changed from \textit{Tachypnea (HP:0002789)} to \textit{Bradypnea (HP:0046507)} and from \textit{Elevated serum creatinine (HP:0003259)} to \textit{Decreased serum creatinine (HP:0012101)}. We hypothesize \textit{Fever} is far more common than \textit{Hypothermia} in the training data, so the model is finetuned with bias towards the highly frequent phenotypes. This may result in the inflation of the scores in Table~\ref{tab:results_mimic_supervised} for \cite{zhang2021selfsupervised} (supervised) which overestimates its numerical reasoning capabilities. 
% It is an open question to assess the impact of such supervision with more annotated data for low frequency phenotypes. 
% On the other hand, above sentences clearly demonstrate that the \textit{NR model can capture phenotypes appearing in different context along with accurate and deep numerical reasoning}.
Based on the observation, we conclude the supervision without additional tailored learning objectives is not sufficient to obtain the numerical reasoning capabilities.

% \red{@Ashwani, add an example NR model fails like FPs and FNs}
However, there are some cases where the NR model fails to produce accurate predictions. For example, in the text - \textit{``Pt still with scant bibasilar crackles. Sat @ 97\% on 2L
NG. Continuing with oral HTN meds and Dig.''}, the model predicts \textit{Abnormal blood oxygen level (HP:0500165)} to \textit{negate} the phenotype \textit{``Sat @ 97\%''} as 97\% is within normal reference range for blood oxygen, i.e., 95\%-100\%. However, the correct phenotype is \textit{Hypoxemia (HP:0012418)} as the patient achieved this normal range using some external oxygen which implies from the phrase \textit{``2L NG''}.  

\subsection{Ablation Studies}
We conduct two ablation studies to probe the benefit of contextualized embeddings and the learning objective for finetuning in Equation \ref{eqn:finetuning_loss}.

% \usepackage{multirow}
\begin{table*}[h]
\centering
\scalebox{0.99}{
\begin{tabular}{|c|c|c|c|c|c|c|}
\hline
\multirow{2}{*}{\textbf{NR Model with}}                                                          & \multicolumn{3}{c|}{\textbf{Exact}}                    & \multicolumn{3}{c|}{\textbf{Generalized}}              \\ \cline{2-7} 
                                                                                              & \textbf{Precision} & \textbf{Recall} & \textbf{F1}     & \textbf{Precision} & \textbf{Recall} & \textbf{F1}     \\ \hline
Keyword based shallow matching                                                                              & \textbf{0.6854}    & 0.2641          & 0.3813          & \textbf{0.7745}    & 0.3449          & 0.4773          \\ \hline
Pretrained contextualized embeddings                                                            & 0.5065             & 0.3758          & 0.4314          & 0.6006             & 0.465           & 0.5241          \\ \hline
\begin{tabular}[c]{@{}c@{}}Finetuned contextualized embeddings \\ (used by the final NR model)\end{tabular} & 0.5176             & \textbf{0.6879} & \textbf{0.5907} & 0.6479             & \textbf{0.7907} & \textbf{0.7122} \\ \hline
\end{tabular}
}
\caption{Ablation studies on the unsupervised test set. Comparison of Numerical Reasoning (NR) model variants using keyword based shallow matching of lexical candidates with numeric entities, pretrained contextualized embeddings and finetuned contextualized embeddings. The finetuned contextualized embeddings substantially outperform other two methods and is incorporated into the final NR model.}
\label{tab:results_ablation_studies}
\end{table*}

\begin{figure*}[!h]
     \centering
     \begin{subfigure}[b]{0.49\textwidth}
         \centering
         \includegraphics[width=\textwidth]{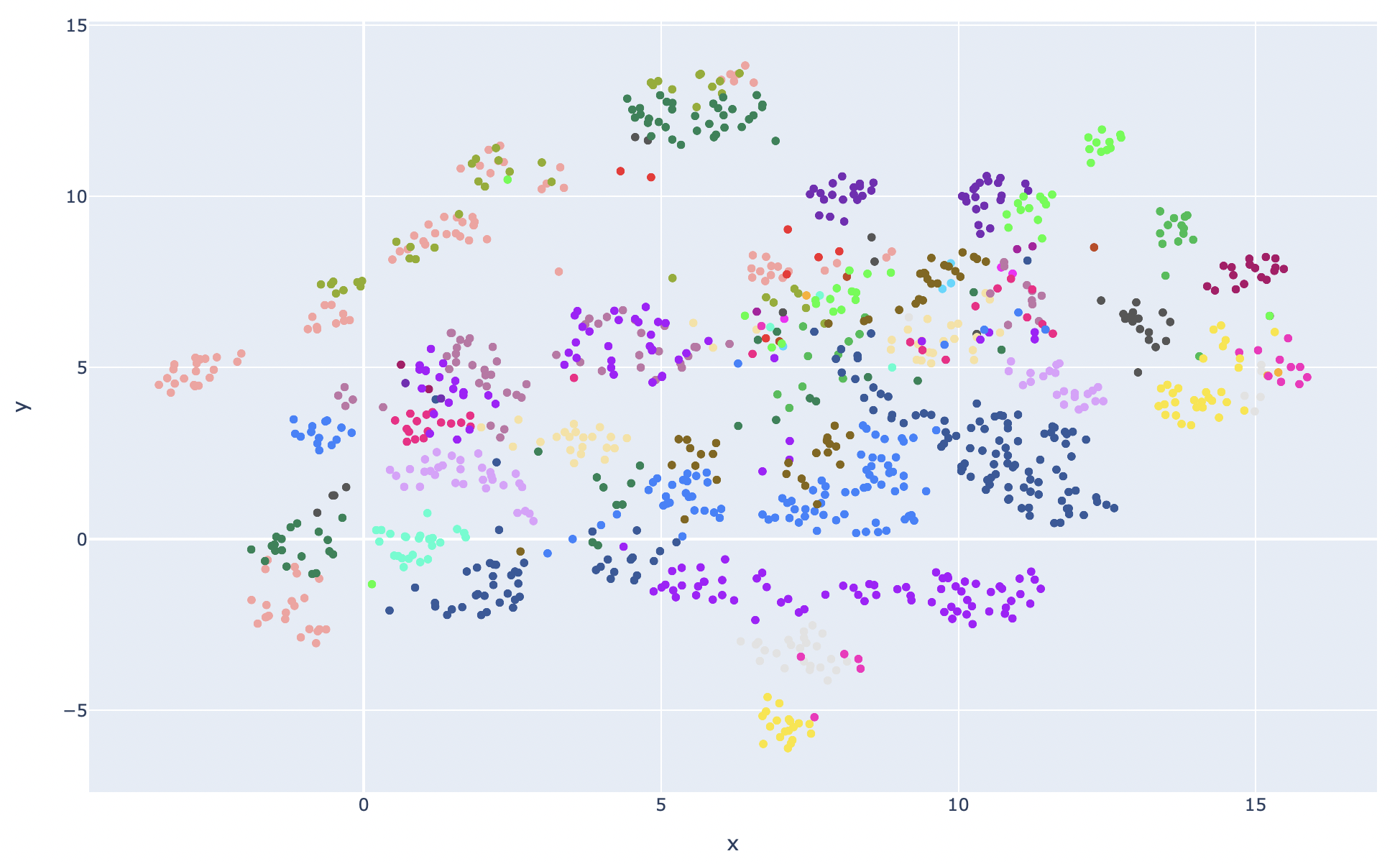}
         \caption{Pretrained contextualized embeddings}
         \label{fig:pretrained_scatterplot}
     \end{subfigure}
     \hfill
     \begin{subfigure}[b]{0.49\textwidth}
         \centering
         \includegraphics[width=\textwidth]{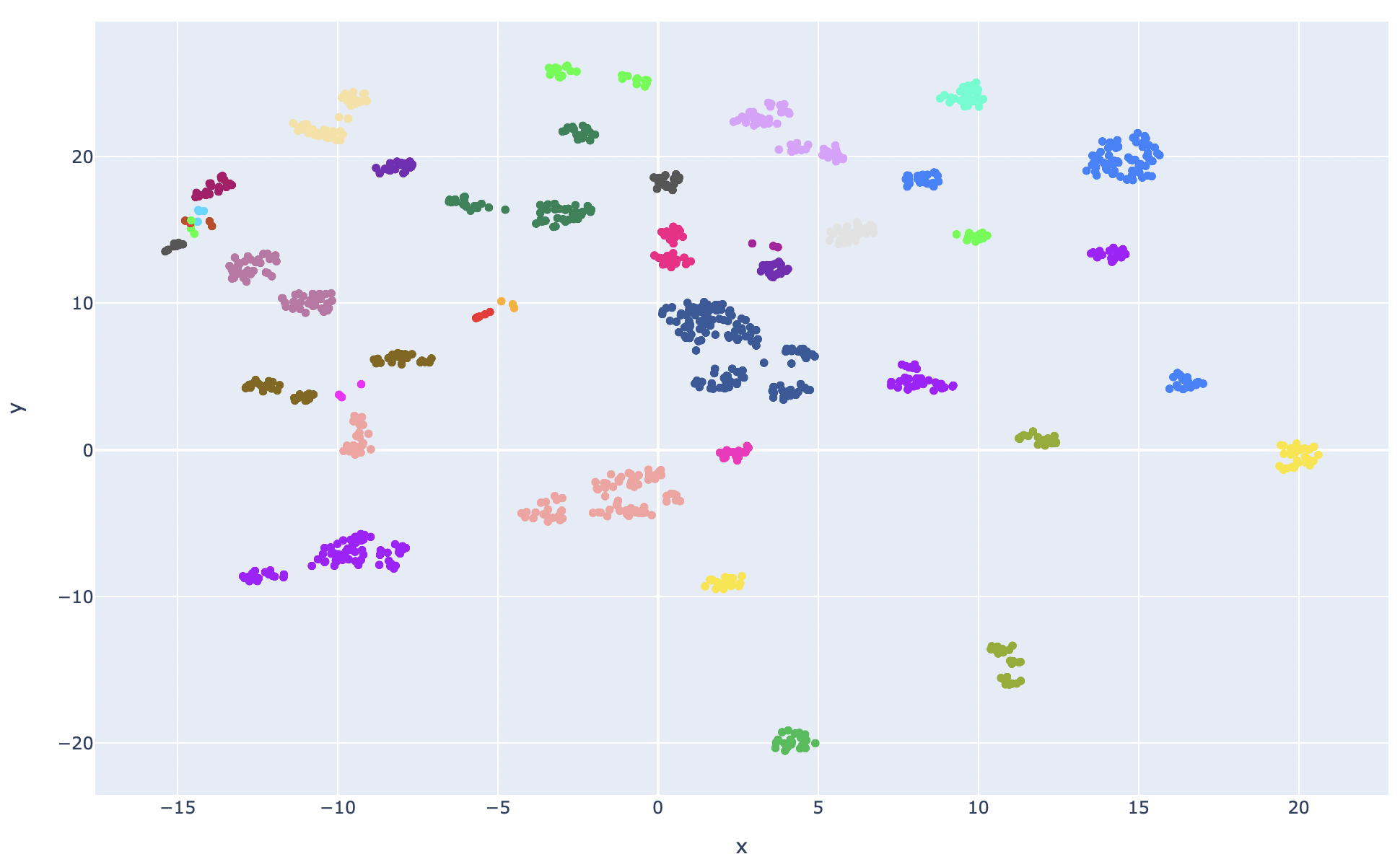}
         \caption{Finetuned contextualized embeddings}
         \label{fig:finetuned_scatterplot}
     \end{subfigure}
     \hfill
     \begin{subfigure}[b]{0.9\textwidth}
         \centering
         \includegraphics[width=\textwidth]{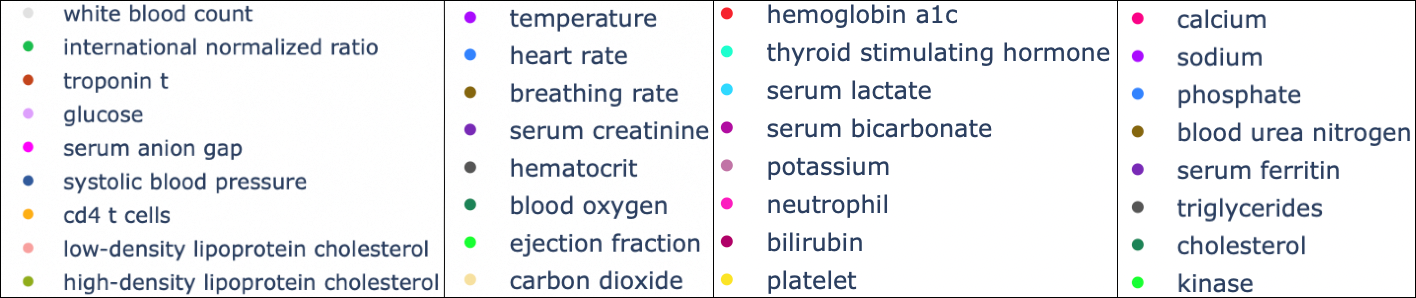}
         \caption{Color codes for numeric entities}
         \label{fig:color_codes_scatterplot}
     \end{subfigure}
    \caption{UMAP visualization of pretrained and finetuned contextualized embeddings of numeric entities and their UMLS synonyms by pretrained and finetuned ClinicalBERT, respectively. Finetuning leads to better differentiation of numeric entities in the semantic space which helps the NR model to identify them with higher accuracy.}
    \label{fig:umap_scatterplot}
\end{figure*}

% \subsubsection{Numeric Entity Extraction using Keyword based Shallow Matching}

To evaluate the usage of contextualized embeddings with cosine similarity to connect lexical candidates with numeric entities as described in Section~\ref{ssec:methodology_relation_prediction}, 
we ablate the contextualized embeddings and instead we use keyword based shallow matching to connect lexical candidates with numeric entities. Table \ref{tab:results_ablation_studies} shows that the ablated method results in significant performance drop, more precisely, in terms of exact Recall from 68.8\% to 26.4\% and F1 from 59.1\% to 38.1\% on unsupervised test set. Therefore, contextualized embeddings is beneficial to capture the semantics of lexical candidates (corresponding to numeric entities) appearing in different contexts.

% \subsubsection{Comparing Pre-trained and Fine-tuned Contextualized Embeddings for Numeric Entity Extraction}
% \label{ssec:results_umap_visualization}

% \red{JZ comment: move the corresponding paragraphs in Methodology, Experiment Setup her}

% \red{JZ comment: one key question to be answered clearly is why the visualisation is needed to show what conclusion? This should be mentioned at the beginning/end of the next paragraph as the opening/conclusive statement.}

In Table \ref{tab:results_ablation_studies}, we also compare the difference between pretrained and finetuned contextualized embeddings. The pretrained embeddings are generated by the pretrained ClinicalBERT model without finetuning and the finetuned embeddings are generated after finetuning ClinicalBERT using Semantic Textual Similarity (STS) Equation \ref{eqn:finetuning_loss} as mentioned in Section \ref{ssec:methodology_sts_contextual_embeddings}.
% After verifying the importance of contextualized embeddings, we ablate fine-tuned embeddings used in the NR model and directly use pre-trained ClinicalBERT embeddings to extract the entities corresponding to the numbers in clinical text using the procedure detailed in Section~\ref{ssec:methodology_relation_prediction}. 
As shown in Table \ref{tab:results_ablation_studies}, the pretrained contextualized embeddings perform poorly with a drop on exact Recall from 68.8\% to 37.6\% and F1 from 59.1\% to 43.1\% on unsupervised test set. 
% Thus, we enhance these embeddings using Semantic Textual Similarity (STS) \cite{DBLP:conf/emnlp/ReimersG19} finetuning objectives. 
For better interpretation, we visualize the pretrained and finetuned contextualized embeddings of numeric entities and their corresponding UMLS synonyms in Figure~\ref{fig:umap_scatterplot} by using Uniform Manifold Approximation and Projection (UMAP) dimensionality reduction \cite{2018arXivUMAP}.
% using the libraries UMAP Learn \cite{mcinnes2018umap-software} and Plotly \cite{plotly} to project the high dimensional embeddings in the 2D space. 
We find that, by the pretrained contextualized embeddings 
% (Figure~\ref{fig:pre-trained_scatterplot}) of only a few numeric entities such as serum creatinine, hematocrit, temperature, etc. with their synonyms occupy a particular region in the space. 
most of the numeric entities are spread out unevenly in the space. For example, the data points for (general) cholesterol, low-density lipoprotein cholesterol, and high-density lipoprotein cholesterol are intermixed. On the other hand, the finetuned contextualized embeddings form well segregated clusters 
% As the model exploits deep contextualized representations, it implicitly clusters together other similar unseen representations which helps in much better generalizability. It powers the model to reason with numeric entities appearing in different context. 
which means it is easier to predict a corresponding numeric entity of lexical candidates (connected with a number) using cosine similarity without collisions. 
% For example, mapping \textit{``Tachypnea''} to \textit{``breathing rate''} instead of \textit{``heart rate''} is straightforward as the fine-tuned contextualized embeddings are not intermixed. Due to the same, the NR model mapped \textit{``breathing rate of 27"} with Tachypnea instead of phenotypes associated with heart rate. 
Overall, it confirms that pretrained contextualized embeddings are not sufficient to connect lexical candidates with numeric entities effectively without the proposed learning objective for finetuning in Equation \ref{eqn:finetuning_loss}.

\section{Conclusions and Future Works}

Numerical reasoning is critical to capture critical phenotypes such as bedside measurement from clinical text. Current state-of-the-art phenotyping models are not designed to reason with numbers, and thus all of them perform poorly in detecting the phenotypes that require numerical reasoning. The proposed unsupervised model shows substantial gains over these models due to its explicit design to reason with numbers by leveraging external knowledge. 
% Further, it can successfully do numerical reasoning in a wide variety of textual context. Upon comparison of our methodology with a supervised baseline model, former surpasses the latter with huge margins using the Recall metric. Further, the supervised model relied on context to determine the phenotypes with numbers without actually doing any reasoning. 
% The phenotypes extracted by the proposed model are also found useful for downstream use cases to identify patients with particular rare diseases  Pulmonary Arterial Hypertension (PAH) and Lupus Nephritis. Also, it helps in finding ICU patients with decompensation. 
The proposed model can be potentially generalized to other biomedical NLP tasks that require numerical reasoning from text. The model can be further extended to consider document level context and dynamic external knowledge base.
% has some limitations such as it is restricted to sentence level context which can be extended to document level context to improve the performance. Also, static external knowledge has some limitations such as strictly defined normal reference range which is vaguely defined in practice. So, incorporating a dynamic knowledge into our methodology is another research direction.   

% \input{sections/others}

\begin{acknowledgement}
We would like to thank Dr. Garima Gupta, Dr. Deepa (M.R.S.H) and Dr. Ashok (M.S.) for helping us create gold-standard phenotype annotation data and validate the external knowledge for numerical reasoning. 
\end{acknowledgement}

% \bibliographystyle{plain}
% \bibliography{ref.bib}
\printbibliography
\end{document}